\documentclass[lettersize,journal]{IEEEtran}
\usepackage{amsmath,amsfonts}
\usepackage{algorithmic}
\usepackage[ruled,vlined]{algorithm2e}
\usepackage{array}
\usepackage[caption=false,font=normalsize,labelfont=sf,textfont=sf]{subfig}
\usepackage{textcomp}
\usepackage{stfloats}
\usepackage{url}
\usepackage{verbatim}
\usepackage{graphicx}
\usepackage{cite}
\usepackage{tikz}
\usepackage{jabbrv}
\usepackage{amssymb}   % 提供 \checkmark
\usepackage{pifont}    % 提供 \xmark
\usepackage{graphicx}
\usepackage{booktabs}
\usepackage{multirow}
\usepackage{xcolor}
\usepackage{tabularx}
\usepackage[font=footnotesize,labelfont=bf,labelsep=colon]{caption}
\usepackage{makecell} 
\usepackage{placeins}

\newcolumntype{?}{!{\vrule width 1pt}}

\newcommand{\bcnum}[1]{%
\tikz[baseline=(char.base)]{
    \node[shape=circle, fill=black, text=white, inner sep=1pt] (char) {#1};}%
}

\begin{document}

\title{Robust Driving Control for Autonomous Vehicles: An Intelligent General-sum Constrained Adversarial Reinforcement Learning Approach}

\author{Junchao Fan, Qi Wei, Ruichen~Zhang, Yang~Lu, Jianhua Wang, Xiaolin Chang, 
and Bo Ai,~\IEEEmembership{Fellow,~IEEE}
        % <-this % stops a space
\IEEEcompsocitemizethanks{
\IEEEcompsocthanksitem Junchao Fan, Qi Wei, and Xiaolin Chang are with the Beijing Key Laboratory of Security and Privacy in Intelligent Transportation, Beijing Jiaotong University, P.R.China. (e-mail: \{23111144, 25121689, xlchang\}@bjtu.edu.cn)
\IEEEcompsocthanksitem Ruichen Zhang is with the College of Computing and Data Science, Nanyang Technological University, Singapore. (e-mail: ruichen.zhang@ntu.edu.sg)
\IEEEcompsocthanksitem Yang Lu is with the School of Computer Science and Technology, Beijing Jiaotong University, P.R.China. (e-mail: yanglu@bjtu.edu.cn)
\IEEEcompsocthanksitem Jianhua Wang is with the College of Computer Science and Technology, Taiyuan University of Technology, P.R.China. (e-mail: wangjianhua02@tyut.edu.cn)
\IEEEcompsocthanksitem Bo Ai is with the School of Electronics and Information Engineering, Beijing Jiaotong University, P.R.China. (e-mail: boai@bjtu.edu.cn)
}}

\markboth{Journal of \LaTeX\ Class Files,~Vol.~14, No.~8, August~2021}%
{Shell \MakeLowercase{\textit{et al.}}: A Sample Article Using IEEEtran.cls for IEEE Journals}

\IEEEpubid{}

\maketitle

% ------------------------------ 
\begin{abstract}
Deep reinforcement learning (DRL) has demonstrated remarkable success in developing autonomous driving policies. 
However, its vulnerability to adversarial attacks remains a critical barrier to real-world deployment.
Although existing robust methods have achieved success, they still suffer from three key issues: (i) these methods are trained against myopic adversarial attacks, limiting their abilities to respond to more strategic threats,  
(ii) they have trouble causing truly safety-critical events (e.g., collisions), but instead often result in minor consequences, 
and (iii) these methods can introduce learning instability and policy drift during training due to the lack of robust constraints. 
To address these issues, we propose Intelligent General-sum Constrained Adversarial Reinforcement Learning (IGCARL), a novel robust autonomous driving approach that consists of a strategic targeted adversary and a robust driving agent. 
The strategic targeted adversary is designed to leverage the temporal decision-making capabilities of DRL to execute strategically coordinated multi-step attacks. In addition, it explicitly focuses on inducing safety-critical events by adopting a general-sum objective. 
The robust driving agent learns by interacting with the adversary to develop a robust autonomous driving policy against adversarial attacks. 
To ensure stable learning in adversarial environments and to mitigate policy drift caused by attacks, the agent is optimized under a constrained formulation.
Extensive experiments show that IGCARL improves the success rate by at least 27.9\% over state-of-the-art methods, demonstrating superior robustness to adversarial attacks and enhancing the safety and reliability of DRL-based autonomous driving.
\end{abstract}

% ------------------------------ 
\section{Introduction}

\IEEEPARstart{T}{he} rapid development of autonomous driving (AD) technology is reshaping transportation systems, with the potential to substantially improve road safety, mitigate traffic congestion, and enhance mobility~\cite{zhao2025survey, sun2025sparsedrive}.
As a prominent AI paradigm, deep reinforcement learning (DRL) has achieved remarkable success in complex sequential decision-making tasks in recent years~\cite{zhang2025embodied, guo2025deepseek}.
Its powerful decision-making capability makes it a promising approach for the realization of AD~\cite{zhaoSurveyRecentAdvancements2024a,al-sharmanSelfLearnedAutonomousDriving2023}.
The recent DRL-based methods have demonstrated their effectiveness across a variety of driving tasks, including highway driving~\cite{zhang2024integration, wang2025highway}, on-ramp merging~\cite{liuReinforcementLearningBasedMultiLane2024, liOnRampMergingHighway2023}, and intersection navigation~\cite{hoelEnsembleQuantileNetworks2023, zhangPredictiveTrajectoryPlanning2023}.

Despite these achievements, the safety and robustness of DRL-based AD policies remain critical factors for their real-world deployment~\cite{huang2024human, zhang2025}. AD is inherently safety-critical with virtually no margin for error. Even small deviations, such as a slight steering overcorrection, can lead to safety-critical events, such as collisions~\cite{wang2025uncertainty, zhaoSurveyRecentAdvancements2024a}. Notably, even well-trained DRL policies often exhibit significant vulnerabilities to adversarial attacks~\cite{ding2023seeing, standenAdversarialMachineLearning2025}. 
Such attacks operate by injecting carefully crafted adversarial perturbations into the agent’s inputs, such as camera images, LiDAR point clouds, or radar signals, to mislead the driving policy of the agent~\cite{fan2024less_is_more}. 
Such vulnerabilities represent a critical bottleneck for deploying DRL in AD, where robustness is not merely desirable but essential.

\begin{figure}[!htbp]
    \centering
    \includegraphics[width=0.5\textwidth, trim=0 10 10 10, clip]{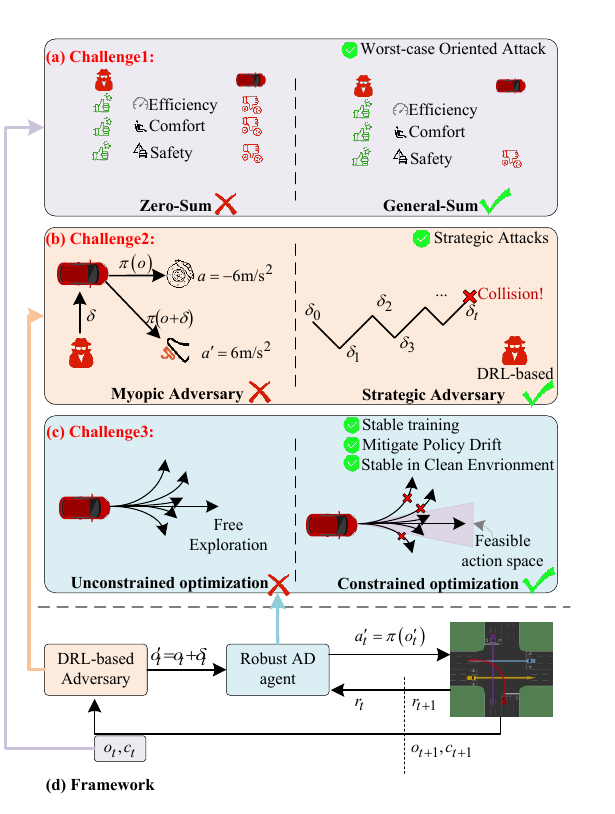} 
    \caption{Overview of IGCARL and its role in addressing key challenges. IGCARL consists of two main components: a strategic targeted adversary designed to find critical vulnerabilities, and a robust driving agent trained to withstand the adversary and produce a robust driving policy.}
    \label{fig:framework}
\end{figure}

Adversarial training has emerged as a widely used defense paradigm for improving the robustness of DRL agents against adversarial perturbations~\cite{li2025enhancing, zhou2024adversarial}. 
The key idea is to enhance robustness by training an agent in competition with an adversary that exploits its vulnerabilities through adversarial perturbations. 
While this technique has shown notable success in strengthening AD policies, existing methods still face at least \textbf{three critical challenges}.
\begin{itemize}
\item \textbf{Oversimplified Zero-Sum Reward Design:} Existing works typically formulate adversarial training as a two-player zero-sum game, where the adversary minimizes the agent’s cumulative reward~\cite{heRobustDecisionMaking2023, wang2025, guoRobustTrainingMultiagent2025}, as shown in Fig.~\ref{fig:framework}(a). 
Although this design is common in other domains, it is not well-suited to AD.
In this context, an adversary’s primary objective is to induce safety-critical events, whereas a driving agent’s reward function must also account for efficiency and comfort.
Consequently, a zero-sum design may bias the adversary toward generating perturbations that affect efficiency or comfort, rather than causing serious safety-critical events~\cite{bai2025rat}. 
Therefore, the challenge lies in designing an adversary with an objective directly coupled to safety-critical events, ensuring that the training process exposes the most critical vulnerabilities.

\item  \textbf{Myopic Adversary Design:} Many existing adversaries adopt a greedy strategy, focusing on immediate gains at each timestep~\cite{heRobustLaneChange2023, wangExplainableDeepAdversarial2024a, heTrustworthyAutonomousDriving2024a}, as shown in Fig.~\ref{fig:framework}(b). 
For instance, the adversary may generate perturbations that maximize the agent’s action deviation at each time step~\cite{heRobustDecisionMaking2023}.
While these perturbations may succeed in specific critical states, this myopic behavior neglects the temporal dynamics intrinsic to driving.
In reality, the effects of an action, such as a slight acceleration change, often manifest with delay, creating opportunities for long-horizon attacks.
A carefully planned sequence of perturbations can compound over time, ultimately inducing severe safety-critical events.
Consequently, the challenge lies in ensuring that agents remain robust against adversaries capable of strategic multi-step attacks, rather than only against myopic, stepwise-optimal perturbations.

\item \textbf{Unconstrained Training Instability:} Existing agents are generally trained under the standard Markov Decision Process (MDP) framework, in which they freely explore and optimize their policies in the presence of adversarial perturbations~\cite{ma2018improved, zhangCATClosedloopAdversarial2023, caiAdversarialStressTest2024, guoRobustTrainingMultiagent2025}, as shown in Fig.~\ref{fig:framework}(c). 
This design overlooks that unconstrained exploration in perturbed dynamics can not only bias the learned policy but also induce training instability. 
The agent may mistake perturbed states as clean states, leading to a failure to learn the safe action distribution in unperturbed environments. 
Moreover, the dynamic nature of adversarial perturbations can further destabilize training, making policy optimization more challenging.
Consequently, the challenge lies in improving robustness against adversarial perturbations while maintaining reliable performance in clean environments.
\end{itemize}

Therefore, we introduce Intelligent General-sum Constrained Adversarial Reinforcement Learning (IGCARL), a novel robust AD approach to address the aforementioned challenges, as shown in Fig.~\ref{fig:framework}(d). 
IGCARL consists of two key components: a strategic targeted adversary designed to find critical vulnerabilities and a robust driving agent trained to withstand the adversary. First, departing from myopic adversaries, the adversary in IGCARL leverages the temporal decision-making and long-term planning capabilities of DRL to generate strategically coordinated multi-step attack sequences. Furthermore, unlike traditional zero-sum formulations, this adversary operates under a general-sum objective, with its reward explicitly decoupled from the agent’s reward. This design allows the adversary to focus directly on inducing safety-critical events. To enhance driving robustness against such an advanced adversary, we train a robust driving agent in the adversary-perturbed environment using adversarial training. Unlike the paradigm of unconstrained optimization, the agent employs constrained policy optimization. Specifically, we impose two key constraints: a collision risk constraint to limit high-risk behaviors and a policy consistency constraint to enforce policy consistency. These constraints jointly ensure stable learning and mitigate policy drift caused by adversarial perturbations. 
To the \textbf{best of our knowledge}, this is the first work to improve the robustness of autonomous driving policies against a stronger type of threat: strategic adversarial attacks that coordinate multiple steps to trigger safety-critical failures.
Our approach offers the following capabilities, while existing approaches only possess some of these, as shown in Table~\ref{tab:related_work}.

\begin{itemize}
  \item \textbf{Strategic Attacks:} The DRL-based adversary can plan strategically coordinated multi-step attack sequences, overcoming the limitations of myopic, single-step attacks.
  \item \textbf{Worst-case Oriented Attacks:} The adversary explicitly focuses on inducing safety-critical events, such as collisions, rather than merely reducing the agent’s performance.
  \item \textbf{Stable Learning:} The robust driving agent is trained using constrained optimization, ensuring stable learning even under changing dynamics caused by adversarial perturbations.
  \item \textbf{Mitigation of Policy Drift:} The policy consistency constraint can prevent the driving policy from overfitting to adversarial perturbations, ensuring reliable performance under both adversarial and clean conditions.
  \item \textbf{Stable in Clean Environments:} The collision risk constraint can ensure the stable performance of the agent even in clean environments.
\end{itemize}

The remainder of this paper proceeds as follows. Section II reviews related work, and Section III details our proposed approach, IGCARL. We then present and analyze experimental results in Section IV, and finally conclude in Section V.

% ------------------------------ 
\section{Related Work}
% 在正文中插入表格
\begin{table*}[!t]
\centering
\caption{Comparison of Related Work and Our Approach}
\label{tab:related_work}
\renewcommand{\arraystretch}{1.2}
\setlength{\tabcolsep}{8pt}
\resizebox{\textwidth}{!}{%
\begin{tabular}{l c c c c c c c c c}
\toprule
\multirow{2}{*}{\makecell{Ref.}} &
\multirow{2}{*}{\makecell{Attack \\ Method}} &
\multirow{2}{*}{\makecell{Defense \\ Method}} &
\multirow{2}{*}{\makecell{Game-Theoretic}} &
\multicolumn{2}{c}{\makecell{Adversary}} &
\multicolumn{4}{c}{\makecell{Agent}} \\
\cmidrule(lr){5-6} \cmidrule(lr){7-10}
% --- 第 2 行表头 ---
% 同样，为下方的表头也强制居中
& & & &
\multicolumn{1}{c}{\makecell{Worst-Oriented}} &
\multicolumn{1}{c}{\makecell{Strategic}} &
\multicolumn{1}{c}{\makecell{Constrained \\ Optimization}} &
\multicolumn{1}{c}{\makecell{Stable \\ Learning}} &
\multicolumn{1}{c}{\makecell{Mitigate \\ Drift}} &
\multicolumn{1}{c}{\makecell{Stable \\ Clean}} \\
\midrule
Fan \emph{et al.}(2024)~\cite{fan2024less_is_more} & \checkmark & \ding{55}  & \ding{55} & \checkmark & \checkmark & \ding{55} & \ding{55} & \ding{55} & \ding{55}  \\
He \emph{et al.}(2023)~\cite{heRobustDecisionMaking2023} & \checkmark & \checkmark  & Zero-sum & \ding{55} & \checkmark & \checkmark & \checkmark & \checkmark & \ding{55} \\
Wang \emph{et al.}(2025)~\cite{wang2025}            & \checkmark & \checkmark  & Zero-sum & \ding{55} & \checkmark & \ding{55} & \ding{55} & \ding{55} & \ding{55} \\
Wang \emph{et al.}(2024)~\cite{wangExplainableDeepAdversarial2024a} & \checkmark & \checkmark  & \ding{55} & \checkmark & \ding{55} & \checkmark & \checkmark & \checkmark & \ding{55} \\
He \emph{et al.}(2024)~\cite{heTrustworthyAutonomousDriving2024a} & \checkmark & \checkmark  & \ding{55} & \checkmark & \ding{55} & \checkmark & \checkmark & \checkmark & \ding{55} \\
\textbf{Ours}                  & \checkmark & \checkmark & General-sum & \checkmark & \checkmark  & \checkmark & \checkmark & \checkmark & \checkmark \\
\bottomrule
\end{tabular}%
}
\begin{flushleft}
\footnotesize
Mitigate Drift: Mitigation of Policy Drift, Stable Clean: Stable Performance in Clean Environments
\end{flushleft}
\end{table*}

Table~\ref{tab:related_work} provides a summarized comparison between our approach and representative prior studies, detailed in Section II.A and II.B.

\subsection{Adversarial Attacks for DRL-based AD}
Adversarial attacks on DRL-based AD policies have raised increasing concerns, as even small perturbations can lead to unsafe or catastrophic behaviors in safety-critical scenarios~\cite{fan2024less_is_more, zhaoSurveyRecentAdvancements2024a}.
In the physical domain, several studies demonstrated attacks such as adversarial patches~\cite{buddareddygariTargetedAttackDeep2022, chahe2024} or manipulations involving surrounding vehicles~\cite{caiAdversarialStressTest2024}, which can mislead perception modules and induce unsafe decisions.

While effective, these attacks are costly to deploy, are sensitive to environmental conditions, and often lack generalizability. 
By contrast, digital-domain attacks that operate directly at the data level or within simulated environments are more flexible, easier to implement, and better suited for systematically probing the fundamental vulnerabilities of DRL-based driving policies.
Among these, the most common are state-space attacks~\cite{fan2024less_is_more, fan2025sharpening}. 
For example, Sun~\emph{et al.}~\cite{sunStealthyEfficientAdversarial2020} demonstrated that by injecting carefully crafted perturbations into the agent’s perceived environmental states, the agent can be misled into taking entirely different actions, ultimately causing policy failure. 
Although these methods contribute to understanding the mechanisms of adversarial attacks, they fall short of providing effective solutions for defending DRL-based AD policies against such threats.

\subsection{Robust DRL-based AD Methods Against Adversarial Attacks}

The deployment of DRL in AD has raised critical safety concerns. 
Due to the black-box nature of DRL policies, their decision-making process is often opaque and difficult to interpret, making it hard to guarantee safety under unseen scenarios~\cite{hu2025}.

To address this issue, some safety-enhancing techniques have been proposed. 
A common line of defense is to incorporate safety checkers that verify whether a candidate action is safe before execution~\cite{heTrustworthyDecisionMakingAutonomous2024, chenAttentionBasedHighwaySafety2024}.
For example, Chen~\emph{et al.}~\cite{chen2020} designed an external safety layer based on a linear biased SVM to filter out unsafe actions from a DRL motion planner. 
Another line of work encourages the agent to learn safety-aware policies by introducing safety constraints during training, integrating them directly into the optimization objective~\cite{al-sharmanSelfLearnedAutonomousDriving2023}.
Additionally, some methods leverage humans or large language models (LLMs) as experts, using interventions or demonstrations to guide agents toward learning safer driving policies~\cite{huangTrustworthyHumanAICollaboration2024a, wuHumanGuidedDeepReinforcement2024, huangSafetyAwareHumanintheLoopReinforcement2024, pangLargeLanguageModel2024}.
For instance, Li~\emph{et al.}~\cite{liEfficientLearningSafe2022} proposed a novel paradigm of expert intervention, enabling human experts to take control in hazardous situations and demonstrate safe driving behaviors to help the agent learn safer driving policies. Sun~\emph{et al.}~\cite{sun2024} proposed an LLM-enhanced RLHF framework that leverages human-controlled agents and feedback in simulation to guide safer AD policies.

However, these methods typically assume a clean environment and overlook the impact of adversarial perturbations on AD policies.
Perturbations injected into sensor inputs can mislead the driving agent, leading to sudden acceleration, lane departures, or abrupt braking, which pose serious safety hazards. 
As a result, a growing body of work investigates defense mechanisms under adversarial environments. 
For instance, He~\emph{et al.}~\cite{heTrustworthyAutonomousDriving2024a} proposed a novel adversarial training method, introducing an adversary that leverages Bayesian optimization and FGSM to find perturbations that aim to maximize collisions . The defender, based on a Robust Constrained Markov Decision Process (RCMDP), learns an optimal and robust driving policy via Lagrangian duality theory. Similarly, Wang~\emph{et al.}~\cite{wangExplainableDeepAdversarial2024a} proposed another adversarial training approach, distinguished by an attacker designed to directly maximize immediate control risk (e.g., throttle and steering). 
While both approaches effectively enhance robustness, the adversaries in them remain inherently myopic. 
These attacks focus on generating an optimal observation perturbation at a single timestep, without considering a longer-term, strategic sequence of correlated perturbations. 
This limitation suggests that existing defenses may be insufficient against sequential attacks that unfold over extended horizons~\cite{fan2024less_is_more}.
Consequently, there is a pressing need to develop robust AD methods that explicitly account for foresighted adversaries, capable of planning temporally correlated perturbations over multiple steps.

Some studies have explored incorporating DRL-based adversaries to enhance attack foresight and proposed corresponding defenses~\cite{heRobustDecisionMaking2023, wang2025}. 
For instance, Guo~\emph{et al.}~\cite{guoRobustTrainingMultiagent2025} proposed a multi-agent state perturbation approach, where adversarial noise is applied to the system states with the objective of minimizing the cumulative rewards of multiple vehicles. 
These methods often formulate the problem as a zero-sum game, where the adversary seeks to minimize the agent’s reward.
However, causing more severe outcomes is more critical in AD. Most driving agents’ reward functions include multiple components, such as traffic efficiency, lane-keeping, passenger comfort, fuel efficiency, and so on. Simply minimizing the cumulative reward does not necessarily lead to critical failures, such as collisions.
To address this, we propose a novel robust AD framework based on a general-sum game paradigm, in which the adversary is explicitly designed to pursue worst-case outcomes. This decoupled objective allows the adversary to focus exclusively on identifying safety-critical vulnerabilities, rather than being distracted by non-critical reward components, thereby creating a more challenging and realistic training environment for enhancing final agent robustness.
% ------------------------------ 
\section{Methodology}
This section first outlines the overall framework of IGCARL in Section III.A. We then introduce the two-player partially observable Markov game formulation that models the interaction between the adversary and the agent in Section III.B. Sections III.C and III.D describe the two core components of IGCARL.

\subsection{Framework}
The overall framework of IGCARL, illustrated in Fig.\ref{fig:framework}(d), consists of two components: \bcnum{1} strategic targeted adversary and \bcnum{2} robust driving agent. 

\begin{itemize}
    \item \textbf{Strategic Targeted Adversary:} We adopt a DRL-based design for the adversary, motivated by two key considerations. First, equipping the adversary with learning capability ensures an ability balance with the driving agent, which is essential for driving the agent to acquire truly robust policies rather than overfitting to fixed attack policies. Second, DRL is well-suited for long-horizon problems, which is particularly valuable in AD tasks, where driving is a sequence of connected decisions. In such scenarios, adversarial perturbations may not cause immediate failures but can accumulate over time, leading to chain-style attacks that cause safety-critical events. Since DRL is reward-driven, we design the adversary’s reward $r^{adv}$ as a sparse cost function tied directly to collision events. This general-sum design decouples the adversary’s objective from the agent’s, thereby encouraging the adversary to focus on generating perturbations that induce safety-critical events. The adversary generates adversarial perturbations in two steps. First, it produces an adversarial action $a^{adv}$ based on the current observation $o$. Then, $a^{adv}$ is used to generate the adversarial perturbation $\delta$ via gradient-based adversarial perturbation generation (PG) methods. Finally, the perturbed observation $o'=o+\delta$ is fed to the agent.
    \item \textbf{Robust Driving Agent:} To strengthen robustness against the adversary, we design a robust driving agent enhances its robustness through adversarial training. Specifically, the agent learns its policy by interacting with the environment in the presence of the adversary. However, standard DRL training is unconstrained and may lead to policy overfitting to adversarial perturbations as well as unstable learning. To address these issues, we optimize the agent’s policy under a constrained framework incorporating two key constraints. (i) The \emph{Collision Risk Constraint} ensures that actions chosen by the agent under clean observations receive low value according to the adversary’s Q-function, thereby preserving the agent’s performance in clean environments. (ii) The \emph{Policy Consistency Constraint} enforces similarity between actions under perturbed and clean observations, reducing policy drift and enhancing stability. Both constraints are incorporated using Lagrangian formulation, enabling a principled trade-off between reward maximization and robustness. This joint optimization allows the agent to maintain high performance while mitigating policy bias and training instability.
\end{itemize}

\subsection{Two-player Partially Observable Markov Game}
Since IGCARL involves two interacting entities, we initially model their interaction as a two-player Markov game. In real-world driving, the agent’s perception is constrained by its sensors, preventing it from fully observing the environmental state. To account for this partial observability, we further extend the formulation to a Two-player Partially Observable Markov Game (TPOMG).
Formally, the TPOMG can be defined as a tuple:
\begin{equation}
    \mathcal{G} = \langle \mathcal{S}, \mathcal{O},\mathcal{O}',\mathcal{A}, \mathcal{A}^{adv}, P, r, r^{adv}, \gamma \rangle,
\end{equation}
where:
\begin{itemize}
    \item $\mathcal{S}$ denotes the true environment state space, which is not fully observable by the agent and the adversary.
    \item $\mathcal{O}$ represents the observation space available to the adversary, consisting of clean observations $o \in \mathcal{O}$.
    \item $\mathcal{O}'$ represents the space of perturbed observations, where $o' \in \mathcal{O}'$ is given by $o' = o + \delta$, corresponding to the agent's actual observations.
    \item $\mathcal{A}$ and $\mathcal{A}^{adv}$ are the action spaces of the agent and adversary, with actions denoted by $a \in \mathcal{A}$ and $a^{adv} \in \mathcal{A}^{adv}$.
    \item $P : \mathcal{S} \times \mathcal{A} \times \mathcal{A}^{adv} \times \mathcal{S} \to [0,1]$ is the environment transition probability.
    \item $r$ and $r^{adv}$ denote the reward functions of the agent and the adversary, respectively.
    \item $\gamma \in [0,1]$ is the discount factor.
    \item $T$ denotes the time horizon, i.e., the total number of time steps in an episode.
\end{itemize}

Unlike the common two-player zero-sum formulation \cite{pintoRobustAdversarialReinforcement2017}, we adopt a general-sum design, where the adversary and the agent have decoupled yet interdependent objectives. The agent and the adversary aim to maximize their respective expected cumulative rewards:
\begin{equation}
\begin{aligned}
\pi^\star &= \arg\max_{\pi} \mathbb{E}\Big[\sum_{t=0}^{T} r_t\Big], \\
\pi^{adv,\star} &= \arg\max_{\pi^{adv}} \mathbb{E}\Big[\sum_{t=0}^{T} r^{adv}_t\Big],
\end{aligned}
\end{equation}
where $\pi^\star$ and $\pi^{adv,\star}$ denote the optimal agent policy and adversary policy, respectively.

Building on the Markov game formulation above, we next present a detailed design of our AD scenario:

\begin{itemize}
    \item \textbf{Observation Space $\mathcal{O}$ and $\mathcal{O}'$:} Instead of explicitly describing the full environment state space $\mathcal{S}$, we focus on the observation spaces, which are the information actually accessible to the agent and adversary during decision-making. Each observation $o \in \mathcal{O}$ comprises two components: (1) the ego vehicle’s speed and heading, and (2) information of the six nearest surrounding vehicles (front, rear, front-left, rear-left, front-right, rear-right), including relative distance, direction, and velocity. The observation dimensions are identical for both the agent and adversary. The key distinction is that the agent observes a perturbed observation $o' \in \mathcal{O}'$, where $o' = o + \delta$ incorporates adversarial perturbations injected by the adversary.
    \item \textbf{Action Space $\mathcal{A}$ and $\mathcal{A}^{adv}$ :} The agent’s action $a \in \mathcal{A}$ and the adversary’s action $a^{adv} \in \mathcal{A}^{adv}$ are defined over the same continuous space. Specifically, each action corresponds to an acceleration value $a, a^{adv} \in [-7.6,\, 7.6]$ m/s\textsuperscript{2} \cite{heTrustworthyAutonomousDriving2024a}.
    \item \textbf{Reward Function $r$ and $r^{adv}$ :} The agent’s reward $r$ balances efficiency and safety. Specifically, it is encouraged to maintain higher speeds while being penalized for safety violations such as collisions:  
    \begin{equation}
    \begin{aligned}
        r &= \frac{v}{v_{\max}} - c(o, a’), \\
        c(o,a’) &=
    \begin{cases}
    1, & \text{if collision}, \\
    0, & \text{otherwise}.
    \end{cases}
    \end{aligned}
    \label{eq:agent_reward}
    \end{equation}
    where $v$ and $v_{\max}$ denote the current and maximum speeds, respectively, and $c(o,a’)$ represents the safety-related cost. In contrast, the adversary’s reward is defined purely based on collisions, i.e., $r^{adv} = c(o, a’).$
\end{itemize}

\subsection{Strategic Targeted Adversary}
The goal of the strategic targeted adversary is to generate observational perturbations that can induce the driving agent to make unsafe actions over multiple steps, leading to safety-critical failures. The objective can be formulated as the following optimization problem:
\begin{equation}
\begin{aligned}
    \max_{\{a^{adv}_{t}\}_{t=0}^{T}} \quad & 
    \mathbb{E} \left[ \sum_{t=0}^{T} \gamma^{t} r^{adv}_t \right] \\
    \text{s.t.} \quad & \delta = PG\!\left(\pi, o, a^{adv}\right), \\
    & o' = o + \delta, \\
    & a = \pi(o), \quad {a}' = \pi(o'), \\
    & \|\delta\|_{\infty} \leq \epsilon.
\end{aligned}
\label{eq:adv_objective}
\end{equation}
where $PG(\cdot)$ denotes the perturbation generation method, and $\|\cdot\|_{\infty}$ represents the $\ell_{\infty}$-norm, which bounds the perturbation magnitude by $\epsilon$.

We adopt the Soft Actor-Critic (SAC) algorithm to train the adversary, motivated by its strong stability and sample efficiency in continuous control tasks ~\cite{huang2024}.
Specifically, SAC maintains a stochastic adversarial policy $\pi^{adv}(a^{adv}|o)$, two soft Q-value networks $Q^{adv}_{1}(o,a^{adv}), Q^{adv}_{2}(o,a^{adv})$, and a value function $V^{adv}(o)$. 
The overall objective is to maximize the entropy-regularized adversarial return:
\begin{equation}
J(\pi^{adv}) = \mathbb{E}_{\tau \sim \pi^{adv}} \Bigg[ 
\sum_{t=0}^{T} \gamma^t \Big( r^{\text{adv}}_t 
+ \alpha \, \mathcal{H}(\pi^{adv}(\cdot|o_t)) \Big) 
\Bigg],
\label{eq:pi_adv_init}
\end{equation}
where $\tau$ denotes trajectories sampled from the replay buffer $\mathcal{D}$. $\alpha$ is the temperature parameter controlling the trade-off between reward and exploration, and
\begin{equation}
\mathcal{H}(\pi^{adv}(\cdot|o)) = - \mathbb{E}_{{a^{adv}}}[\log \pi^{adv}(a^{adv}|o)] 
\end{equation}
denotes the entropy of the adversarial policy, which encourages exploration and prevents premature convergence.

Following the SAC update scheme, the Q-value networks are optimized by minimizing the Bellman residual:
\begin{equation}
J(Q^{adv}_i) = 
\mathbb{E}_{\tau \sim \mathcal{D}}
\Big[ \tfrac{1}{2} \big(Q^{adv}_i(o_t,a^{adv}_t) - \hat{Q}^{adv}(o_t,a^{adv}_t) \big)^2 \Big],
\label{eq:Q_adv}
\end{equation}
with the target value $\hat{Q}^{adv}(o_t,a^{adv}_t)$ defined as
\begin{equation}
\begin{aligned}
\hat{Q}^{adv}(o_t,a^{adv}_t) &= r^{adv}(o_t,a^{adv}_t) \\
&\quad + \gamma \, \mathbb{E}_{o_{t+1}} \Big[ 
\min_{i=1,2} Q^{adv}_i(o_{t+1},a^{adv}_{t+1}) \\
&\qquad - \alpha \log \pi^{adv}(a^{adv}_{t+1}|o_{t+1}) 
\Big].
\end{aligned}
\end{equation}

Next, the adversarial policy can be reformulated from Eq.~\eqref{eq:pi_adv_init} into a tractable surrogate loss:
\begin{equation}
J(\pi^{adv}) = 
\mathbb{E}_{\tau \sim \mathcal{D}} \Big[
\alpha \log \pi^{adv}(a^{adv}|o) - Q^{adv}(o,a^{adv})
\Big],
\label{eq:pi_adv}
\end{equation}
where $Q^{adv} = \min(Q^{adv}_{1}, Q^{adv}_{2})$.  

Finally, the value function is trained to
minimize the squared residual error:
\begin{equation}
\begin{aligned}
J(V^{adv}) &= 
\mathbb{E}_{s \sim \mathcal{D}} \Bigg[
\tfrac{1}{2} \Big( V^{adv}(o) \;-\; 
\mathbb{E}_{a^{adv}} \big[ Q^{adv}(o,a^{adv}) \\
&\qquad - \alpha \log \pi^{adv}(a^{adv}|o) \big] 
\Big)^2 \Bigg]
\end{aligned}
\label{eq:V_adv}
\end{equation}

Given a clean observation $o$, the adversarial policy $\pi^{adv}$ first outputs an adversarial action:
\begin{equation}
a^{adv} \sim \pi^{adv}(o).
\end{equation}
This adversarial action guides the generation of the adversarial perturbation $\delta$ through the adversarial loss:
\begin{equation}
J_{PG}(\delta) = \big\| a^{adv} - \pi(o+\delta) \big\|^2.
\label{eq:pgloss}
\end{equation}
We adopt the Basic Iterative Method (BIM) ~\cite{kurakin2018adversarial} as our perturbation generation function $\text{PG}(\cdot)$. At each iteration, the perturbation is updated as
\begin{equation}
\delta \gets clip\Big(
\delta + \alpha_\delta \, sign(\nabla_\delta {J}_{PG}(\delta), - \epsilon , + \epsilon
\Big),
\label{eq:bim}
\end{equation}
where $\alpha_\delta$ is the step size, and $clip(\cdot)$ ensures $\delta$ remains within $[-\epsilon, \epsilon]$.

\subsection{Robust Driving Agent}
The objective of the robust driving agent is to enhance policy robustness against adversarial perturbations through adversarial training. 
To prevent overfitting to $o'$ and ensure stable learning process, the agent optimizes its policy under the following constrained formulation:
\begin{equation}
\begin{aligned}
\max_{\{a'_t\}^T_{t=1}} \quad &  
\mathbb{E}\Big[\sum_{t=1}^{T}\gamma^{t}r_t\Big] \\
\text{s.t.} 
\quad &
C1:\mathbb{E}\big[\min_{i\in \{1, 2\}}Q^{adv}_i\!\big(o,\,a\big)\big]\le \epsilon_1, \\
\quad & 
C2:\mathbb{E}\big[\big\| \pi(o)-\pi(o') \big\|^2\big]\le\epsilon_2,
\label{eq:robust}
\end{aligned}
\end{equation}
where $\epsilon_1$ and $\epsilon_2$ are thresholds. $C1$ and $C2$ denote the Collision Risk Constraint and Policy Consistency Constraint, respectively.

\begin{figure*}[htbp]
    \centering
    \includegraphics[width=1\textwidth,trim=10 20 10 20,clip]{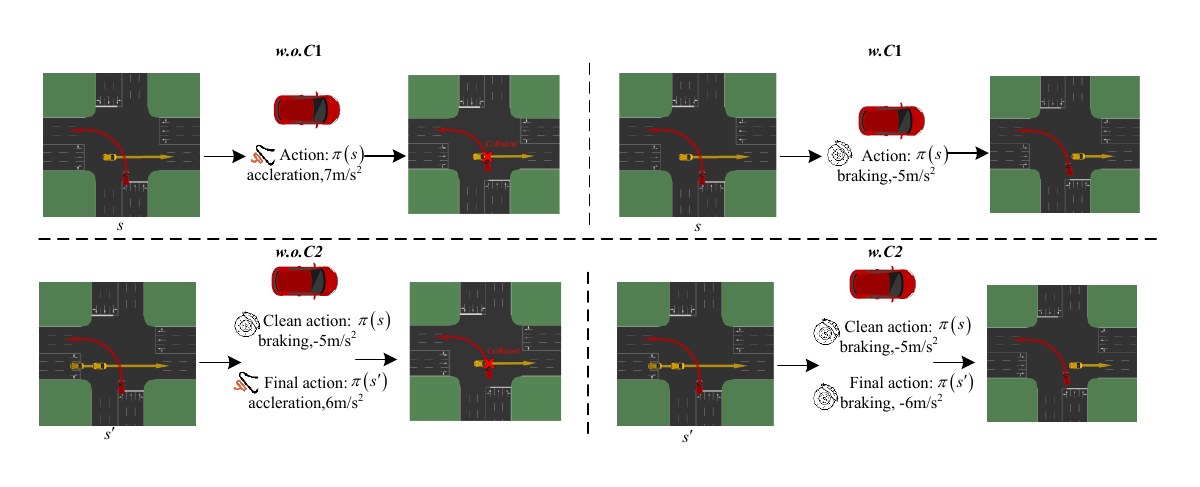} 
    \caption{Qualitative analysis of constraints (C1 and C2) in an unprotected left turn scenario. The top row shows C1: Without C1 (left), the agent selects a hazardous acceleration leading to a collision. With C1 (right), the agent brakes safely. The bottom row shows C2: without C2 (left), the safe braking is overridden by a final acceleration; with C2 (right), the braking is preserved, preventing a collision.}
    \label{fig:constraint}
\end{figure*}

C1 enforces low adversary Q-values for the agent's actions in clean observations, where $Q^{adv}(o,a)$ estimates the potential cost since $r^{adv} = c(o,a)$. 
Without C1, policy shifts caused by training in adversarial environments may lead the agent to misinterpret clean observations, selecting actions with high adversary Q-values.
For instance, the agent may accelerate left when a yellow vehicle approaches, as illustrated in Fig. \ref{fig:constraint}.
With C1, the agent chooses actions with low adversary Q-values, ensuring robust and safe behavior under unperturbed conditions.

C2 constrains the agent’s actions under perturbed observations to stay close to those under clean observations, reducing policy drift caused by adversarial perturbations \cite{heTrustworthyAutonomousDriving2024a}. 
Without C2, the action distributions before and after perturbation can differ significantly.
For instance, the agent may decelerate under clean observations but accelerate under perturbed observations, potentially causing collisions, as illustrated in Fig.~\ref{fig:constraint}.
With C2, the agent maintains small action deviations (e.g., from $-5\,\text{m/s}^2$ to $-6\,\text{m/s}^2$), enhancing robustness against adversarial perturbations.

To address the constrained problem in Eq.~\eqref{eq:robust}, we introduce Lagrange multipliers $\lambda_1, \lambda_2 \ge 0$ and form the primal--dual objective:
\begin{equation}
\mathcal{L}\!\left(\pi,\lambda_1,\lambda_2\right)
=
\mathbb{E}\Big[\sum_{t=1}^{T}\gamma^{t}r(o_t,\tilde{a}_t)
-\lambda_1\big(C1-\epsilon_1\big)
-\lambda_2\big(C2-\epsilon_2\big)\Big].
\label{eq:lagrangian}
\end{equation}
The corresponding dual problem is:
\begin{equation}
\min_{\lambda_1,\lambda_2}\;\max_{\pi}\;\mathcal{L}(\pi,\lambda_1,\lambda_2).
\label{eq:lagdual}
\end{equation}
We optimize Eq.~\eqref{eq:lagdual} via alternating updates: the policy parameters $\pi$ are updated to maximize the Lagrangian, while the multipliers $\lambda_1$ and $\lambda_2$ are adjusted to penalize constraint violations. 
Formally, the optimal parameters satisfy
\begin{equation}
\begin{aligned}
\pi^\star &=\arg \max_{\theta} \mathcal{L}(\pi,\lambda_1,\lambda_2), \\
\lambda_1^\star, \lambda_2^\star &= \arg \min_{\lambda_1,\lambda_2 \ge 0} \; \mathcal{L}(\pi^\star,\lambda_1,\lambda_2).
\end{aligned}
\label{eq:lagrangian_obj}
\end{equation}

We adopt an actor-critic framework to optimize Eq.~\eqref{eq:lagrangian_obj}, which consists of one actor network $\pi$ and two critic networks $Q_{1},Q_{2}$. The use of two critics mitigates the estimation bias in policy updates.

The critic parameters are updated by minimizing the squared Bellman residual:
\begin{equation}
J(Q_i)=
\mathbb{E}\left[\Big(Q_{i}(o_t,a_t)-\big(r+\gamma\min_{j\in \{1, 2\}}Q_{j}(o_{t+1},a_{t+1})\big)\Big)^2\right].
\label{eq:Q_agent}
\end{equation}  

The actor network $\pi$ can be optimized via the primal--dual objective \eqref{eq:lagrangian}. Concretely, the policy gradient takes the form
\begin{equation}
J(\pi)=
\mathbb{E}\!\left[
\min_{j \in \{1,2\}} Q_{j}\!\big(o,a\big)
-\lambda_1(C1-\epsilon_1)
-\lambda_2(C2-\epsilon_2)\right].
\label{eq:pi_agent}
\end{equation}

Finally, the multipliers $\lambda_1,\lambda_2$ are updated by dual gradient ascent:
\begin{equation}
\lambda_k \leftarrow \big[\lambda_k + \alpha_\lambda (C_k-\epsilon_k)\big]_+,\quad k\in\{1,2\},
\label{eq:lambda}
\end{equation}
with $\alpha_\lambda$ the dual stepsize, and $[\cdot]_+ = \max(\cdot, 0)$ ensures non-negativity.

% ------------------------------ 
\section{Experiments}
We validate the performance of IGCARL on the SUMO platform ~\cite{lopezMicroscopicTrafficSimulation2018}. Sections IV.A–D detail the environment setting, baselines, evaluation metrics, and training configurations. We then present and analyze the experimental results to address the following research questions

\begin{itemize}
    \item \textbf{RQ1:} Does IGCARL outperform the baselines and state-of-the-art methods? (See Section IV.E)
    \item \textbf{RQ2:} Does IGCARL maintain robustness under other types of perturbations? (See Section IV.F)
    \item \textbf{RQ3:} Does IGCARL maintain robust performance when deployed in environments different from the training conditions? (See Section IV.G)
\end{itemize}

\subsection{Environment Setting}
We evaluate our approach in a representative high-risk scenario: the unprotected left turn. 
This maneuver is widely regarded as one of the most challenging in urban driving, due to its elevated collision risk and complex interactions with oncoming traffic~\cite{zhaoUnprotectedLeftTurnBehavior2023, al2023self}. 
In the SUMO simulation, vehicles follow the LC2013 lane-changing model ~\cite{erdmann2015sumo}, with a maximum speed of 15 m/s and maximum acceleration/deceleration of 7.6 m/s\textsuperscript{2} to reflect realistic dynamics~\cite{heTrustworthyAutonomousDriving2024a}. Traffic density is controlled by the per-second vehicle arrival probability $p$, which is set to $0.5$ unless otherwise specified.

\subsection{Baselines}
We select three categories of DRL-based AD methods as baselines: 

\begin{itemize}
    \item \textbf{Vanilla RL:} We implement vanilla RL methods based on three representative algorithms: PPO\cite{schulmanProximalPolicyOptimization2017a}, SAC~\cite{haarnojaSoftActorCriticOffPolicy2018}, and TD3~\cite{fujimotoAddressingFunctionApproximation2018}. PPO is a SOTA on-policy algorithm~\cite{fanEnergyConstrainedSafePath2024}, while SAC and TD3 represent SOTA off-policy algorithms~\cite{zhou2024t,dong2025,liu2025resource}. This selection spans both major branches of reinforcement learning, enabling a comprehensive and balanced comparison.
    \item \textbf{Safe RL:} Safe RL methods enhance the action safety of agents by incorporating mechanisms such as safety checkers and safety constraints. For this category, we select SAC\_Lag~\cite{ha2021learning} and FNI~\cite{heFearNeuroInspiredReinforcementLearning2024a} as representative methods.
    \item \textbf{Robust RL:} Robust RL methods aim to strengthen policy robustness by employing techniques such as adversarial training to defend against adversarial perturbations. In this category, we adopt DARRL~\cite{heTrustworthyAutonomousDriving2024a} as the SOTA method.

\end{itemize}

\subsection{Metrics}
We evaluate the performance of AD agents using three widely adopted metrics: task success rate (SR), collision rate (CR), and driving efficiency (DE). 
All metrics are computed over 200 evaluation episodes to ensure statistical reliability.
In our scenario, an episode is considered successful if the ego vehicle reaches the target lane without any collisions. 
DE is quantified as the average speed over the entire episode. 
\subsection{Training Details}
All experiments are conducted on a platform equipped with an Intel(R) Xeon(R) Gold 6230 CPU and an NVIDIA GeForce RTX 4090 GPU. IGCARL and all baseline methods are implemented in PyTorch. Table~\ref{tab:parameters} summarizes the key parameters, including environment settings (Env), DRL hyperparameters (DRL), and PG method parameters (PG). Parameter choices are guided by prior studies\cite{heTrustworthyAutonomousDriving2024a,fan2024less_is_more} and extensive empirical tuning to achieve optimal performance.
\begin{table}[!t]
\centering
\caption{Summary of Experimental Parameters}
\label{tab:parameters}
\renewcommand{\arraystretch}{1.2}
\setlength{\tabcolsep}{8pt}
\begin{tabular}{l l l}
\toprule
\textbf{Category} & \textbf{Parameter} & \textbf{Value} \\
\midrule
\multirow{5}{*}{Env} 
& Max speed $v_{max}$ & 15 m/s \\
& Max acceleration / deceleration & [-7.6,7.6] m/s\textsuperscript{2} \\
& Traffic density $p$ & 0.5 \\
& Number of episodes & 3000 \\
& Max steps per episode & 30 \\
\midrule
\multirow{8}{*}{DRL} 
& Actor learning rate & 1e-4 \\
& Critic learning rate & 1e-3 \\
& Tmperature parameter $\alpha$ & 0.1 \\
& Dual stepsize $\alpha_{\lambda}$ & 5e-5 \\
& Discount factor $\gamma$ & 0.99 \\
& Constraint thresholds $\epsilon_{1},\epsilon_{2}$ & 0.01 \\
& Batch size & 128 \\
& Replay Buffer size & 1e6 \\
\midrule
\multirow{3}{*}{PG} 
& Perturbation magnitude $\epsilon$ & 0.01,0.03,0.05 \\
& Number of iterations & 50 \\
& Update step size $\alpha_\delta$ & $\epsilon/50$ \\
\bottomrule
\end{tabular}
\end{table}

\subsection{Performance Evaluation}
\begin{table*}[h]
\centering
\caption{Performance Comparison Under Different $\epsilon$ With and Without Attacks.}
\label{tab:performance}
\begin{tabular}{l l l l l ? l l ? l l}
\toprule
Condition & Metric & PPO & SAC & TD3 & SAC\_Lag & FNI & DARRL & IGCARL \\
\midrule
\multirow{3}{*}{w.o attacks} 
& SR & $\textcolor{red}{\mathbf{100.00 \pm 0.00}}$ & $\textcolor{red}{\mathbf{100.00 \pm 0.00}}$ & $74.40 \pm 15.11$ & $\textcolor{red}{\mathbf{100.00 \pm 0.00}}$ & $\textcolor{red}{\mathbf{100.00 \pm 0.00}}$ & $98.50 \pm 3.35$ & $\textcolor{red}{\mathbf{100.00 \pm 0.00}}$ \\
& CR & $\textcolor{red}{\mathbf{0.00 \pm 0.00}}$ & $\textcolor{red}{\mathbf{0.00 \pm 0.00}}$ & $22.30 \pm 14.28$ & $\textcolor{red}{\mathbf{0.00 \pm 0.00}}$ & $\textcolor{red}{\mathbf{0.00 \pm 0.00}}$ & $1.50 \pm 3.35$ & $\textcolor{red}{\mathbf{0.00 \pm 0.00}}$ \\
& MS & $13.05 \pm 0.08$ & $\textcolor{red}{\mathbf{13.11 \pm 0.08}}$ & $13.04 \pm 0.17$ & $13.12 \pm 0.04$ & $12.76 \pm 0.08$ & $12.78 \pm 0.13$ & $12.57 \pm 0.25$ \\
\midrule
\multirow{3}{*}{$\epsilon=0.01$} 
& SR & $41.30 \pm 31.47$ & $\textcolor{red}{\mathbf{99.20 \pm 0.45}}$ & $57.00 \pm 22.14$ & $98.50 \pm 0.79$ & $90.40 \pm 2.51$ & $97.40 \pm 2.48$ & $96.00 \pm 3.98$ \\
& CR & $58.70 \pm 31.47$ & $\textcolor{red}{\mathbf{0.60 \pm 0.55}}$ & $34.40 \pm 24.42$ & $1.40 \pm 0.65$ & $9.60 \pm 2.51$ & $2.50 \pm 2.29$ & $4.00 \pm 3.98$ \\
& MS & $13.17 \pm 0.29$ & $13.09 \pm 0.14$ & $12.39 \pm 0.33$ & $\textcolor{red}{\mathbf{13.11 \pm 0.22}}$ & $12.99 \pm 0.03$ & $12.70 \pm 0.11$ & $12.53 \pm 0.47$ \\
\midrule
\multirow{3}{*}{$\epsilon=0.03$} 
& SR & $3.50 \pm 0.82$ & $35.38 \pm 19.76$ & $3.38 \pm 1.65$ & $16.75 \pm 1.32$ & $50.25 \pm 0.65$ & $57.25 \pm 23.30$ & $\textcolor{red}{\mathbf{74.50 \pm 1.32}}$ \\
& CR & $96.50 \pm 0.82$ & $64.62 \pm 19.76$ & $45.25 \pm 2.99$ & $83.25 \pm 1.32$ & $49.75 \pm 0.65$ & $42.75 \pm 23.30$ & $\textcolor{red}{\mathbf{25.50 \pm 1.32}}$ \\
& MS & $13.62 \pm 0.02$ & $\textcolor{red}{\mathbf{13.58 \pm 0.03}}$ & $8.20 \pm 0.03$ & $13.56 \pm 0.01$ & $13.45 \pm 0.03$ & $12.67 \pm 0.66$ & $12.71 \pm 0.26$ \\
\midrule
\multirow{3}{*}{$\epsilon=0.05$} 
& SR & $1.00 \pm 0.50$ & $6.50 \pm 6.95$ & $5.33 \pm 0.58$ & $4.83 \pm 3.69$ & $7.00 \pm 0.50$ & $68.17 \pm 2.31$ & $\textcolor{red}{\mathbf{87.17 \pm 1.76}}$ \\
& CR & $99.00 \pm 0.50$ & $93.50 \pm 6.95$ & $76.50 \pm 31.18$ & $92.33 \pm 8.55$ & $93.00 \pm 0.50$ & $31.83 \pm 2.31$ & $\textcolor{red}{\mathbf{12.83 \pm 1.76}}$ \\
& MS & $13.67 \pm 0.04$ & $13.36 \pm 0.53$ & $11.79 \pm 2.90$ & $12.33 \pm 2.25$ & $13.63 \pm 0.04$ & $13.08 \pm 0.28$ & $\textcolor{red}{\mathbf{12.79 \pm 0.32}}$ \\
\bottomrule
\end{tabular}
\end{table*}
To quantitatively evaluate the robustness of IGCARL against adversarial perturbations, we consider three cases with $\epsilon \in \{0.01, 0.03, 0.05\}$ and compare its performance with the baseline methods. The detailed results are reported in Table~\ref{tab:performance}. We summarize the key findings as follows.

\textbf{Collapse of Vanilla RL:} In the absence of attacks, PPO, SAC, and TD3 achieve success rates above 90\%. Once adversarial perturbations are introduced, their performance drops dramatically. When $\epsilon$ reaches $0.05$, the success rate of all three methods falls below 10\%, indicating complete failure.

\textbf{Limited gains from Safe RL:} SAC\_Lag and FNI, which were originally designed to improve safety, show stronger resistance than vanilla RL. When $\epsilon = 0.01$, they can still maintain success rates above 90\%. However, their advantage diminishes quickly and eventually collapses as $\epsilon$ increases. These results indicate that these methods offer limited protection against adversarial perturbations, as they do not explicitly consider such attacks.

\textbf{Superior robustness of IGCARL:} IGCARL consistently outperforms the strongest baseline, DARRL. IGCARL maintains a 100\% success rate in the w.o.attacks scenario, and demonstrates greater stability compared to DARRL.
When $\epsilon = 0.03$ and $\epsilon = 0.05$, IGCARL achieves success rates that are 30.1\% and 27.9\% higher than DARRL, respectively. In terms of driving efficiency, IGCARL is just 2.5\% slower than SAC\_Lag, which remains the fastest method under the \emph{w.o. attack} condition.
These results confirm that adversarial perturbations have little influence on IGCARL’s decision-making process.

\textbf{Overall insight:} The results confirm that IGCARL effectively learns robust driving policies by combining adversarial training with a strategic targeted adversary. Additionally, by employing a Lagrangian dual formulation to regularize policy updates, IGCARL suppresses policy shifts under adversarial perturbations while maintaining high performance in the no-attack scenario.

\subsection{Robustness Evaluation}
To comprehensively characterize the robustness boundaries and potential weaknesses of IGCARL and all basline methods, we design three complementary experiments that examine policy behaviors from three perspectives: \textbf{dynamic decision-making process}, \textbf{gradient-based perturbations}, and \textbf{random perturbations}.

\subsubsection{Dynamic decision-making process} 
\begin{figure*}[!htbp]
    \centering
    \includegraphics[width=1\textwidth,trim=10 20 10 20,clip]{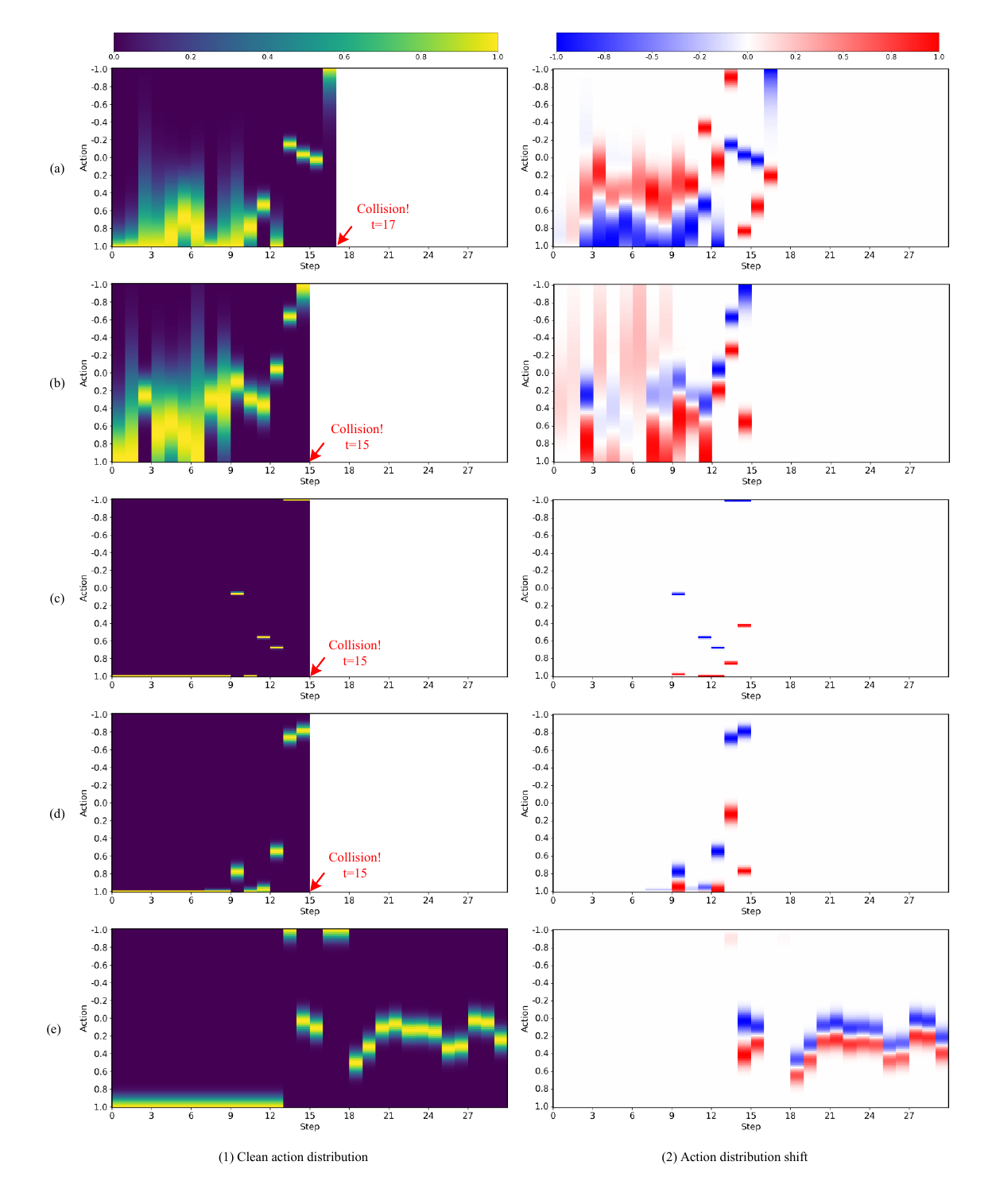} 
    \caption{Visualization of action distributions under clean and adversarial environments. 
    Subfigures (a)–(e) correspond to SAC, SAC\_Lag, FNI, DARRL, and IGCARL, respectively. Column (1) shows the action distribution under clean observations, while Column (2) shows the shift in the action distribution caused by adversarial perturbations, denoted as $\Delta = \pi(o') - \pi(o)$}
    \label{fig:visual}
\end{figure*}

Fig.~\ref{fig:visual} illustrates the impact of our adversary on different methods under $\epsilon=0.05$ in a test episode with a fixed environment seed. 
Specifically, Fig.~\ref{fig:visual}(1) shows the per-timestep action distributions of each method in the clean setting, while Fig.~\ref{fig:visual}(2) presents the corresponding action distribution shift caused by adversarial perturbations, defined as $\Delta = \pi(o') - \pi(o)$. 
Subplots (a)–(e) correspond to SAC, SAC\_Lag, FNI, DARRL, and IGCARL, respectively. 
Among the vanilla baselines, we only report SAC since it consistently outperforms PPO and TD3, making it the most representative choice. 

As shown in Fig.~\ref{fig:visual}(a1)-(b1), SAC and SAC\_Lag exhibit highly variable action distributions during the first 10 steps, corresponding to the stage before entering the intersection.
This inherent policy instability makes them immediately vulnerable to adversarial attacks, resulting in significant policy shifts shown in Fig.~\ref{fig:visual}(a2)-(b2). 
By contrast, FNI, DARRL, and IGCARL exhibit highly concentrated action distributions during this stage, reflecting decisive and consistent policies that experience minimal shift under attack. 

All baselines except IGCARL experience collisions around steps 15 to 20, the stage of traversing the intersection. 
These collisions are preceded by multiple steps of adversarial attacks. 
While the per-step shifts in the action distribution may not have an immediate effect, their cumulative effect gradually pushes the agent’s trajectory off the safe manifold, ultimately resulting in collisions.

In contrast, IGCARL demonstrates superior robustness throughout the episode. Its policy remains virtually unchanged during the first 14 steps. Even in later stages beyond step 20, only minor deviations occur. This indicates that IGCARL can overcome policy shifts caused by adversarial perturbations and demonstrates robustness against strategically coordinated multi-step adversarial attacks.

\subsubsection{Gradient-based perturbations} 
\begin{figure*}[!htbp]
    \centering
    \includegraphics[width=1\textwidth]{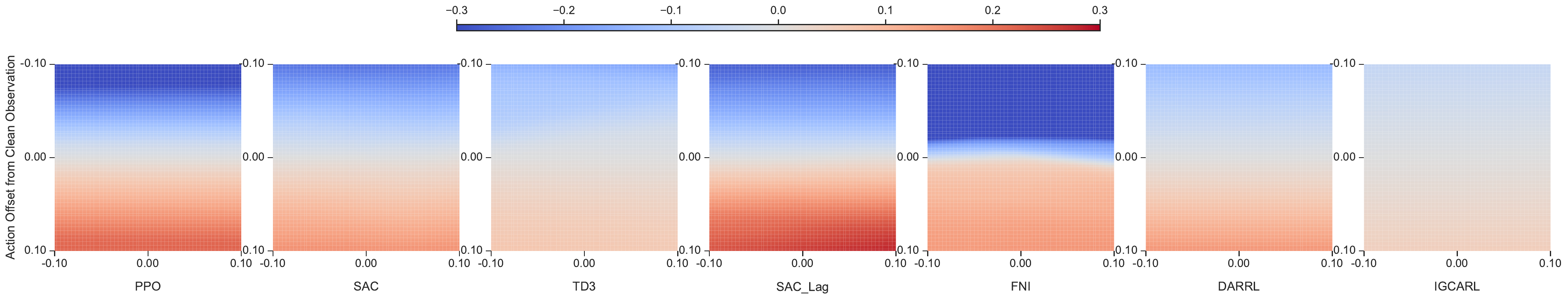} 
    \caption{Sensitivity analysis of policies under gradient-based perturbations. The y-axis denotes the perturbation along the action gradient direction, while the x-axis represents the perturbation along the direction orthogonal to the action gradient.}
    \label{fig:gradient_perturbations}
\end{figure*}
The previous experiments demonstrate that IGCARL effectively withstands multi-step adversarial perturbations generated by the adversary. 
However, such attacks are shaped by the adversary and may not fully reveal the inherent sensitivity of the learned policy. 
To further assess the local stability of the policy beyond adversary-driven scenarios, we design a gradient-based myopic adversarial attack. 
This attack perturbs the agent’s observations along two directions: the gradient of the action with respect to the observation, and an orthogonal direction. 
Specifically, the perturbation is applied using the action gradient $\nabla_o \mu(o)$ and an orthogonal unit vector $u$, formulated as
\begin{align}
& o' = o + \beta_1 \frac{\nabla_o \mu(o)}{\|\nabla_o \mu(o)\|} + \beta_2 u,
u \perp \nabla_o \mu(o), \\
& \|\beta_1 \frac{\nabla_o \mu(o)}{\|\nabla_o \mu(o)\|} + \beta_2 u\| \le \epsilon_{max},
\end{align}
where $\mu(o)$ represents the mean action output by the policy $\pi$. $\beta_1$ and $\beta_2$ are coefficients controlling the perturbation magnitude along the gradient and orthogonal directions, respectively. This ensures that the combined perturbation is bounded by $\epsilon_{\text{max}}$.
The max perturbation magnitude is set to $\epsilon_{max}=0.1$. 
Our analysis is based on five clean observations sampled from the entire task episode.
The x-axis denotes the perturbation along the orthogonal direction, while the y-axis denotes the perturbation along the gradient direction.

As shown in Fig.~\ref{fig:gradient_perturbations}, the action offsets vary more prominently along the gradient direction, demonstrating that perturbations along this direction can effectively induce policy shifts. Compared to other methods, IGCARL exhibits remarkable robustness, with action offsets remaining within $±0.1$ in all cases, indicating that it also ensures local stability of the policy. Notably, TD3 appears to be the most stable among all baseline methods, likely due to its deterministic policy, which ensures that minor perturbations cause only negligible changes in its action. However, its poor performance against our adversary highlights that local smoothness does not guarantee overall robustness.

\subsubsection{Random perturbations} 
\begin{figure}[!htbp]
    \centering
    \includegraphics[width=0.4\textwidth]{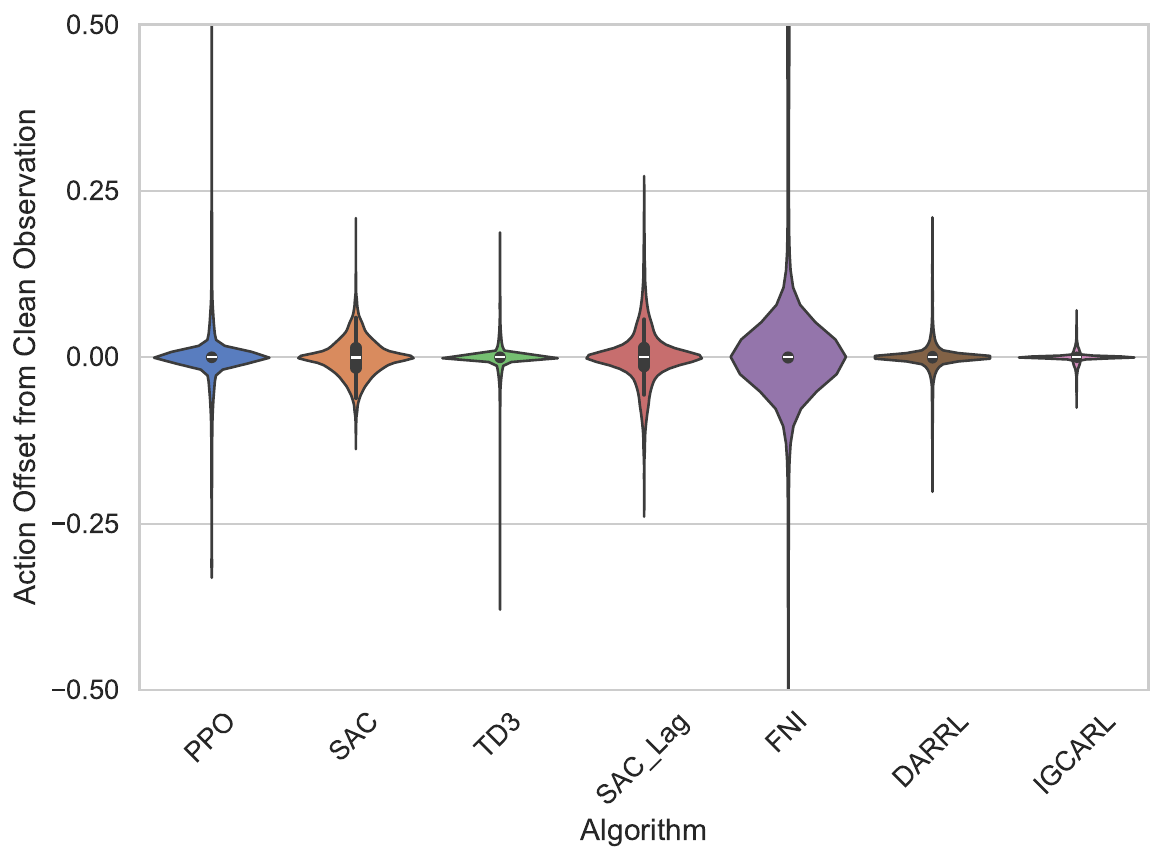} 
    \caption{Action offset under random noise perturbations with $\epsilon=0.05$.}
    \label{fig:actiondistribution}
\end{figure}
\begin{figure*}[h]
    \centering
    \includegraphics[width=1\textwidth]{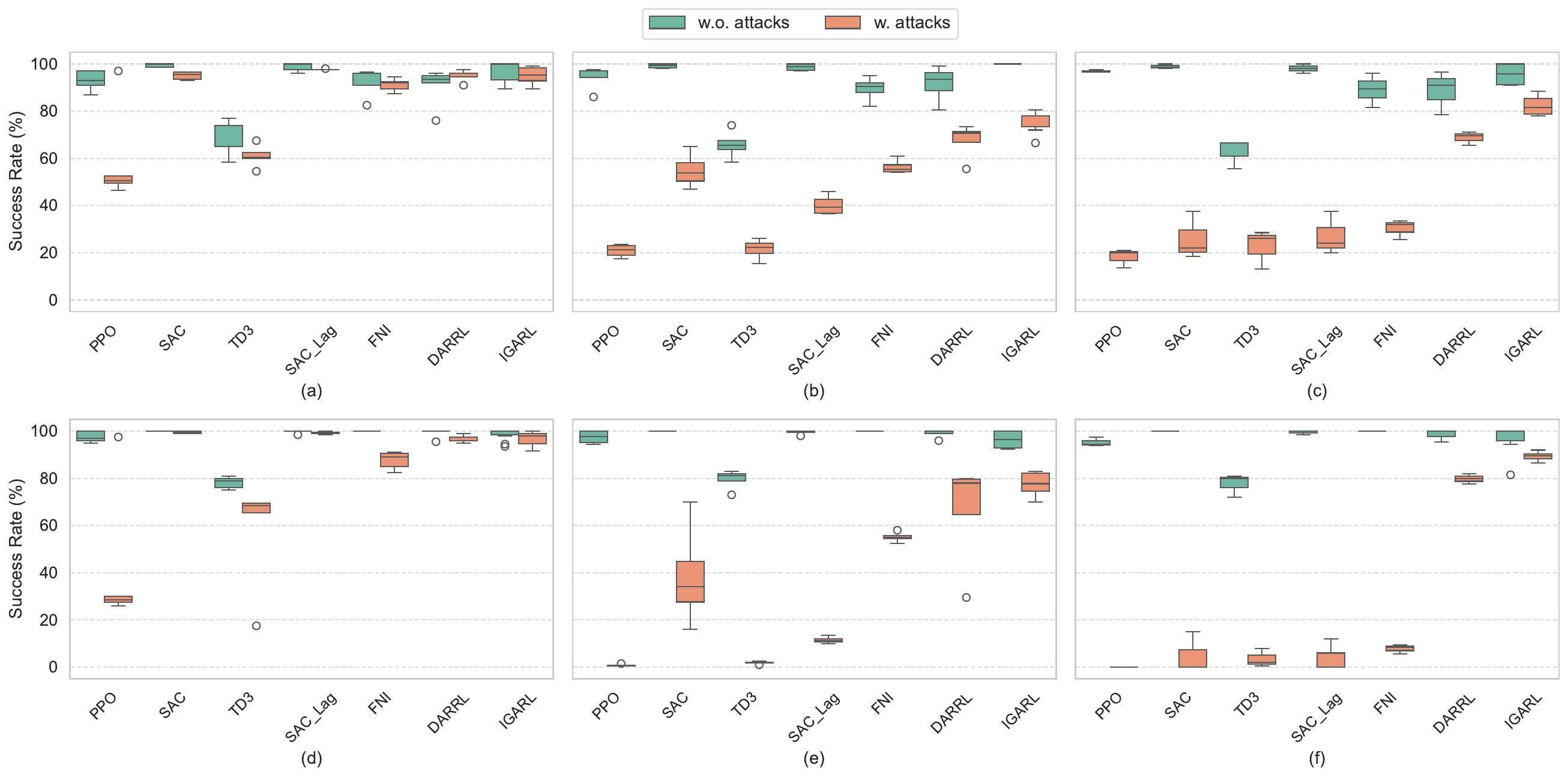} 
    \caption{Performance comparison under different traffic densities. (a)–(c) show results with $\epsilon=0.01, 0.03,$ and $0.05$ in Env1 ($p=0.03$), while (d)–(f) present the corresponding results in Env2 ($p=0.07$).}
    \label{fig:diffdensity}
\end{figure*}
We further evaluate the robustness of the methods by introducing random noise. 
This design is motivated by several considerations: 
(i) random noise can simulate small, unpredictable perturbations in real-world environments, such as sensor measurement errors, providing insight into the policy's stability under non-ideal conditions; 
(ii) random noise does not rely on policy information and thus tests the method's robustness to general, non-adversarial perturbations.
In this experiment, the perturbed observations are sampled uniformly from a sphere of radius $\epsilon=0.05$.
Fig.~\ref{fig:actiondistribution} illustrates the offset of actions induced by perturbations relative to the unperturbed actions.
It can be observed that IGCARL exhibits remarkably strong robustness to random perturbations, with action deviations not exceeding 0.15. 
In contrast, DARRL exhibits a few instances of action deviations exceeding 0.2, indicating that its robustness to policy perturbations does not generalize well to random perturbations.
Although random perturbations may not cause severe consequences, these moderate action deviations can still lead to unpleasant riding experience and undermine confidence in the reliability of AD policies.

\subsection{Generalization Evaluation}
An effective AD policy should not only perform well in its training environment but also generalize effectively to novel, unseen scenarios. 
Traffic density is a critical and dynamically changing factor in real-world applications that directly impacts the complexity and safety of driving decisions. 
Therefore, we evaluate IGCARL in test environments with two different traffic densities $p=[0.03,0.07]$, denoted as Env1 and Env2, respectively.
From Fig.~\ref{fig:diffdensity}(a) and Fig.~\ref{fig:diffdensity}(d), we observe that under low perturbation magnitude ($\epsilon=0.01$), all baseline methods adapt well to changes in traffic density, showing no significant performance degradation. 
However, when the perturbation magnitude increases to $\epsilon=0.03$, PPO, SAC, and TD3 exhibit performance drop in Env2 compared to Env1, as higher traffic density increases the likelihood of collisions. 
With further increases in perturbation magnitude, FNI also suffers a substantial decline, with its SR in Env2 dropping from around 80\% to below 10\%. 
In contrast, IGCARL and DARRL maintain relatively stable performance, with IGCARL consistently outperforming DARRL under attack. 
The gap becomes most pronounced at $\epsilon=0.05$, where IGCARL achieves about 10\% higher SR than DARRL.
In summary, these results demonstrate that IGCARL can still remain robust under varying traffic densities, confirming its strong generalization capability.

% ------------------------------
\section{Conclusion}
In this paper, we have proposed IGCARL, a novel approach for enhancing the robustness of DRL-based autonomous driving policies against adversarial attacks.
IGCARL consists of two key components: a strategic targeted adversary and a robust driving agent.
The strategic targeted adversary leverages the temporal decision-making capabilities of DRL to execute strategically coordinated multi-step adversarial attacks. Furthermore, by adopting a general-sum objective, it explicitly focuses on inducing safety-critical events, such as collisions.
The robust driving agent employs constrained optimization to mitigate policy drift caused by adversarial perturbations during training while ensuring stable learning.
Empirical results indicate that IGCARL exhibits significantly improved robustness against policy-based, gradient-based, and random perturbations. 
These findings underscore the potential of IGCARL to improve the safety and reliability of autonomous driving systems.

% ------------------------------
\FloatBarrier
\bibliographystyle{jabbrv_IEEEtran}
\bibliography{main}

% ------------------------------

\end{document}